# Machine Learning, Clustering and Polymorphy

Stephen José Hanson

Bell Communications Research

*Abstract*

This paper reports a machine induction program (WITT) which attempts to model human categorization. Properties of categories that human subjects are sensitive to include, best or prototypical members, relative contrasts between putative categories, and polymorphy (niether necessary or sufficient features). This approach represents an alternative to traditional Artificial Intelligence (AI) approaches to generalization and conceptual clustering which tend to focus on necessary and sufficient feature rules, equivalence classes, and search and match algorithms. The present approach is shown to be more consistent with human categorization while potentially including results produced by more traditional clustering schemes. Applications of this categorization approach are also discussed in the domains of Expert systems and Information retrieval.

## Introduction

Most current work done in Artificial Intelligence on machine learning and conceptual clustering--and for that matter most generalization schemes that have been proposed in AI—typically rest on five **false** premises:

> *(1) that necessary and sufficient feature lists must be central to the categorization engine;*
>
> *(2) that categories are equivalence classes;*
>
> *(3) that polymorphy (neither necessary or sufficient features) rules are either uninteresting or noise;*
>
> *(4) that probability measures are antagonistic to symbolic manipulation;*
>
> *(5) and that the top four assumptions are consistent with human categorization data.*

In contrast, psychological results in the categorization literature are inconsistent with each of the five premises above. People do not seem to try to form categories by determining the necessary and sufficient set of "defining features" (Michalski, 1980) for a set of objects.[1] Rather people seem to form relative "contrasts" between categories; that is, people tend to minimize "variance" within clusters while maximizing

---

1. The distinction drawn here is somewhat subtle and does not imply that people and animals do not have or know about categories that possess necessary and sufficient features, that would imply that people could not use common features which is contrary to intuition. However, there are many possible mechanisms for achieving necessary & sufficient categories, as exemplified in the "contrast approach" advocated below. Nonetheless, such categories for the present approach are a special case rather then a central purpose of the categorization engine.

"variance" between clusters (Rosch & Lloyd, 1978; Smith & Medin, 1981 ). People also tend to have best or prototypical members of a category as opposed to equivalence classes (Homa, 1978; Posner & Keele,1968). Many categories that people use (perhaps all natural categories) have all or at least some members that possess neither necessary nor sufficient features and can best described by a polymorphy rule ("m features out of n", m<n) (Dennis, Hampton & Lea, 1973; Smith & Medin 1981, chapter 4,"probabilistic features"; see Figure 1 for an example of polymorphous categories). Finally, although the evidence is mixed, it seems clear that people are able to use "likelihood" estimates (in a Bayesian sense) that members belong together, that is, they seem to be sensitive to the density of members in a local feature space or equivalently to the *information* content[2] (cf. Orloci, 1969) of the members lying in a given feature space neighborhood. This is in contrast to typical clustering methods that use similarity measures that are derived from such probability or information measures since the distribution of features across objects must represent the first level of data input to any categorizer.

*Rationale*

We motivate the following conceptual clustering scheme on the psychological categorization literature reviewed above. This allows us to make clear the specific assumptions underlying the notion of conceptual clustering and the nature of the properties of psychological categories:

> *(1) Categories arise as "contrasts" between one another, in other words, categorization is relative to the existing field of other putative categories.*

> *(2) Categories have a distribution of members, some more representative, some less. Furthermore there tends to be one best or a set of best members or an abstracted "member" (prototype) described in terms of features that can be used to represent the entire category.*

> *(3) Categories tend to possess members that are polymorphous, especially when there is either natural variation in members or when the category is supported by a rational or causal account of the underlying semantic relations in the category.*

> *(4) Categories can be represented by the likelihood that members tend to belong together. In contrast to similarity measures, which can be shown to be arbitrary, information measures are distribution free, sensitive to the probability of features within entities, and related to the raw input of features.*

> *(5) Categories and categorization should be motivated by psychological research, because categorization is a basic process that underlies many artificial intelligence domains including expert systems, natural language processing, semantic networks, as well as information retrieval. Both the human comprehensibility of the categorization and its match to human performance can be argued to be crucial to the realization of progress in each of these areas.*

A "concept" will be defined as having four properties: (1) an identity which can be described in terms of the feature space; (2) prototypical or best members; (3) layers of boundaries that introduce more polymorphy into the category; (4) and a relative tension or contrast between a given concept and any other concept in the field.

---

2. Information is used in a technical sense to refer to Shannon (1948) Information (H). It is used as a measure of the "likelihood" that entities belong together or have some intrinsic contiguity in a feature space.

The approach described in this paper is distinct from statistical clustering, which has been primarily motivated to provide different views of the same data or to explore data by using arbitrary similarity metrics and rules for group membership (cf. Everitt, 1977). Widely used statistical methods for clustering typically admit three kinds of metrics and three kinds of group membership rules, although there are clustering packages (e.g., CLUSTAN 1B, Wishart,1969) that effectively have over 400 different ways to do clustering analysis of the same data!

In contrast, conceptual clustering is an attempt to derive the categories that would be most consistent with a semantic or structural interpretation in which the members could have been described. It is a "weak" approach that uses very little prior information about the nominal nature of the categories ("isa" or "kindof" or property lists); nonetheless, nothing so far would preclude the addition of further knowledge about the category or the input entities. Basically, this approach attempts to use the known psychological properties of categories and find the most likely representation of the given entities based on input features. Finally, note at this point in the development of this approach that feature selection is not attempted.

Because similarity can be defined in so many ways and is psychologically controversial,[3] it is reasonable to assume that people can deal with the raw data input prior to any presumed psychological similarity transforms. We assume that they can form probability estimates within a feature space that might be described in everyday reasoning contexts (e.g. "the probability that the cafeteria is closed for coffee"); that these probability estimates can be used to form local contrasts between potential categories (cf. Tversky, 1977); and that the category structure includes prototypes, polymorphy, and a tension between overgeneralization and identity of the category relative to all other categories as they are forming.

*Program Flow*

The present Conceptual Clustering algorithm (WITT[4]) attempts to automatically cluster a set of objects with a given level of polymorphy which is predefined by a set of parameters. The general strategy of WITT could be described as a generate-and-test algorithm with escalation over category formation states. There are three such states as shown in Figure 2: object hunting, protoseed hunting, and prototype merging, each with its own goals and procedures. WITT cycles through each state until it hits an impasse, that is, it finds it cannot precede with the present goal. A new state is then invoked with a corresponding set of new goals, and WITT attempts a new set of hypotheses.

For example, given a set of objects defined on a multi-valued feature space, WITT first attempts to form local estimates of highly dense regions by using an *information loss* metric;[5] these regions are then assigned to a "protoseed".

At each cycle WITT begins to test whether it is possible to add members to each protoseed without affecting the "identity" of each putative category. If "object

---

3. For example, different similarity measures can be used to recover many different structures within the same data, furthermore, non-dimensional stimuli like words do not seem to be best described by models that are appropriate for dimensional stimuli like colors.

4. Named for the philosopher Ludwig Wittgenstein who argued persuasively for "family resemblance" and polymorphy as the basis for categorization and language. The classic example of this problem is the nature of the category "game".

   WITT is implemented in Franz-Lisp on a Pyramid and has also been ported to a Symbolics 3600.

5. Such measures are usually defined in terms of independence within the feature space: $H(o_1,o_2) - H(o_1) - H(o_2)$ These measures are also used in WITT to establish TRANSMISSION between objects in the the feature space. This can be shown to be similar to a category validity approach which maximizes the conditional probability of an category given a feature (Gluck & Corter, 1985).

hunting" fails, WITT tries a new hypothesis and enters the "protoseed hunting" state in which new dense regions are found and assigned to new protoseeds. If the identities of all present protoseeds are maintained or improved, then the protoseed is instantiated and "object hunting" state is re-entered.

If, on the other hand, the new protoseed fails, WITT attempts one more hypothesis, that protoseeds are too close together to gain improvement. At this point, "protoseed merging" is attempted and identities of the protoseeds are again checked. If successful, WITT returns to "object hunting "; otherwise WITT quits, announcing a categorization as well as the objects still unclustered, and provides a natural language description (in terms of existing feature labels input) of the prevailing concept field.

At the end of the process three properties of the category are guaranteed: (1) at least one best member is identified for each category; (2) at least the level of polymorphy that was initially requested exists in the concept space; (3) a relative contrast between all categories is chosen to maximize "identity" within a category and minimize overgeneralization between categories.

*Implications for Some Applications in AI and Information Retrieval*

*Categorization forms the basis for most kinds of applications in AI and information retrieval. The present view should imply that there are fundamental problems in using categorization results that are motivated from necessary and sufficient feature lists (either conjunctive or disjunctive).*

### Expert Systems

Clancy (1984) has shown that expert system knowledge representation can be described as a classification problem. But the classification that experts do can also be generally shown not to be based on necessary and sufficient features.[6] The canonical structures and behavior of expert systems, according to Clancy are: "...identifiable phases of data abstraction, heuristic mapping onto a hierarchy of pre-enumerated solutions and refinement within this hierarchy. In short these programs do what is commonly called classification." The actual heuristic match that determines the assignment of a given set of data to a particular class is what is of interest here. These classes typically consist of generalizations from the input data, and simulation of what a human expert might do when mapping a set of conditions to an another abstraction that corresponds to an appropriate action.

These mappings tend to be rules of thumb, numerical weightings or even "probabilistic" in nature; generalizations that are true some of the time or have a certain likelihood of being true. That is, some input data is considered ideal or prototypical; other features of input data are acceptable; while still others might be marginal. In other words, in any such expert system there must be a miss-classification rate that defines a certain rate of false alarms and misses, presumably acceptable to the expert.

These kinds of mappings that exist in expert systems violate the necessary and sufficient rule and the equivalence class rule, as well as potentially admit polymorphous rules or worse probabilistic ones. Consequently, expert system knowledge would benefit from a categorization scheme like the present one which can generate classifications based on multivalued features that should more readily capture expert's classifications since they are based on human performance in classification tasks. It should be noted that such work is beginning to appear in expert systems literature; for example, Butler and Corter (1985) used a traditional statistical clustering technique to model computer program trouble-shooters using structural analysis, and subsequently used the results of the clustering to build an expert system for this domain used now by Boeing Computer Services.

---

6. although there are such experts, REX, for example, that do have a simple decision trees of necessary and sufficient features it is questionable that actual statistical reasoning can be captured in such a simple scheme.

*Information Retrieval*

Information retrieval is another area which might benefit from clustering that bears similarity to human categorization, since people presumably use their concepts about a field of study to attempt to retrieve information in it. The majority of information retrieval techniques involve a specification of a conjunction or disjunction of conjunctions (disjunctive normal form) of attributes or keywords in order to define a document for retrieval. Other techniques are probabilistic and involve correlations between keywords specified by a user and their occurrence within a set of documents. All such methods tend not to respect the category structure that the user brings to the system.

As mentioned earlier, specific areas of study or research are not likely to be defined by a conjunctive concept. That is, it is probably not the case that say, "altruism in redwing sparrows" or "automaticity in learning" as two areas of research respectively from biology or psychology have necessary and sufficient attributes or boolean combinations of keywords[7] that allow admittance into the category but not into other categories. In fact, as shown in an experiment below, information retrieval categories tend to have best or prototypical members; they tend to admit polymorphy (neither necessary or sufficient attributes); and they are distinct only in the context of other existing concepts.

Specifying information retrieval categories may be most compatible with users concepts by indication of the "center" or best member of the retrieval category ("a seminal paper in the area....") and then specifying other members through a polymorphous rule, that is, a coordinate indexing (a and b or a and c or b and c; actually best stated as a polymorphous 2 out of 3 rule on the the set of attributes (a,b,c)). In the next section we attempt to look at the some of the issues in a pilot experiment on information retrieval categories using experts.

*Some Results*

The first issue to examine is the comprehensibility of categories that are based on necessary and sufficient feature lists and represent the traditional assumptions underlying conceptual clustering approaches in AI. The second experiment attempts to generate the same categories that experts use when describing documents from their area of research.

*Human Studies*

The first question we wanted to investigate is how human subjects do with some of the categorization results from more traditional AI conjunctive clustering algorithms. Claims have been made that results from such clusterers are more comprehensible because they use criteria that attempt to find a feature list description that is parsimonious and provides disjoint cover (e.g. Michalski's "LEF" criteria). Yet no simple empirical test of the comprehensibility of these categories has been attempted.

An example of results from a Conjunctive Clusterer is shown in Figure 3. Each toy train has a set of features which includes the length of the train, car shape, contents, size, etc... about 10 features with 3-4 values per feature. Michalski's conjunctive clusterer (Michalski & Stepp, 1983; CLUSTER/PAF) describes the 5 cars at the top of the figure as a conjunction of short car and a closed top and the 5 bottom cars as those trains that have either two cars or trains that have a car with a jagged top. These two "concepts" provide complete disjoint cover across the categories. Thus, these feature descriptions are necessary and sufficient for group inclusion. Are these categories also comprehensible to people?

---

7. There is, of course, always a disjunctive normal form that would allow specification of all the members of the category without probabilistic interpretation, one is an M out of N rule (polymorphy) and the other is the trivial case of a disjunction of N conjunctions for N members.

Subjects were asked to do 5 unique sorts of the 10 cars into two piles (to ensure that all features were considered). In a second task subjects were asked to guess the correct rule for category inclusion by looking first at 2 trains each from one or the other category, then 4 trains, 6 trains and so on until all 10 trains were present with their group membership indicated.

There were 13 subjects, 5 of whom were either from a mathematics or computer science research group at Bell Communications Research. The results were clear and quite unambiguous: none of the 13 subjects were able to sort the stimuli into the conjunctive clusters found by the Michalski clustering algorithm. One subject (a computer scientist) came close to stating the rule in the second task. All other subjects settled on categorizations that allowed polymorphy and had indicated some best train within their set. Most subjects were surprised at the disjunction used to describe one of the train categories and none could reasonably describe the category that each disjoint cover cluster represented.

### WITT Studies

This study compared expert subject's categorizations with the output of WITT using the same input data. Five psychologists (two Faculty and three graduate students) from Princeton University Psychology Department were asked to sort seven preselected psychological abstracts into as many categories as they preferred and to provide the experimenter with a rationale for each category when they were done. Each subject was also asked to indicate how each of their categories combined in order to provide a complete hierarchical description.

Each abstract consisted of the title, author and between two to eight keywords (shown in Figure 4). Subjects were allowed to continue to sort until they were satisfied with their categories. The seven abstracts were selected to have one necessary and sufficient feature for potentially four overlapping categories (e.g. "visual search" for abstracts 4 and 6; "problem solving" for abstracts 1 and 2; and "words" for abstracts 3 and 6).

The results of the sortings, shown in Figure 5, are represented as trees indicating the order of combination of the abstracts from the bottom of the tree to the top. Four out of five subjects had a similar number of categories with similar group membership, while three subjects who had chosen three categories also had identified six out of seven members in the same groups and finally, two of subjects had completely identical categories. The number of categories is indicated by the dashed line cutting the tree and the descriptions subjects gave for each category are indicated below its members.

WITT was given the keywords and unique descriptors from each title that were not already part of the keyword list. This resulted in 12 unique binary variables for each abstract. An intermediate value of polymorphy (.4 cohesion and .2 distinctiveness) was found which produced a categorization that was identical to two of subjects. WITT's tree is shown in the same Figure (Figure 5) as the subject's data in bottom right corner. WITT lables the branches of tree with keywords/descriptors which form polymorphous rules for category admittance and indicates necessary and sufficient features where it finds them.

Even in this simple preliminary experiment the complexity of the result is surprising. Neither from necessary or sufficient features nor common features (those contributing to a correlation) would the general categorization results have been well predicted. However, WITT, using contrasts between categories and allowing polymorphy into each category was able to capture the general results and descriptions that each subject offered as rationale for their categories.

In sum, preliminary results so far suggest that WITT is more compatible with and more comprehensible to human subjects sorting stimuli within the same domain than conceptual clustering that forces equivalence classes and "disjoint cover" (e.g. Michalski's Conceptual Clusterers) when the results of both methods are compared. Human subjects have great difficulty with many of the toy problems that exhibit disjoint cover and are more likely to assign members according to a polymorphous rule.

Acknowledgment

M. Bauer coded most of WITT, G. Collier coded and helped design an early prototype categorizer on which some of WITT was based and finally this manuscript benefited from commments made by R. Allen and D. Walker.

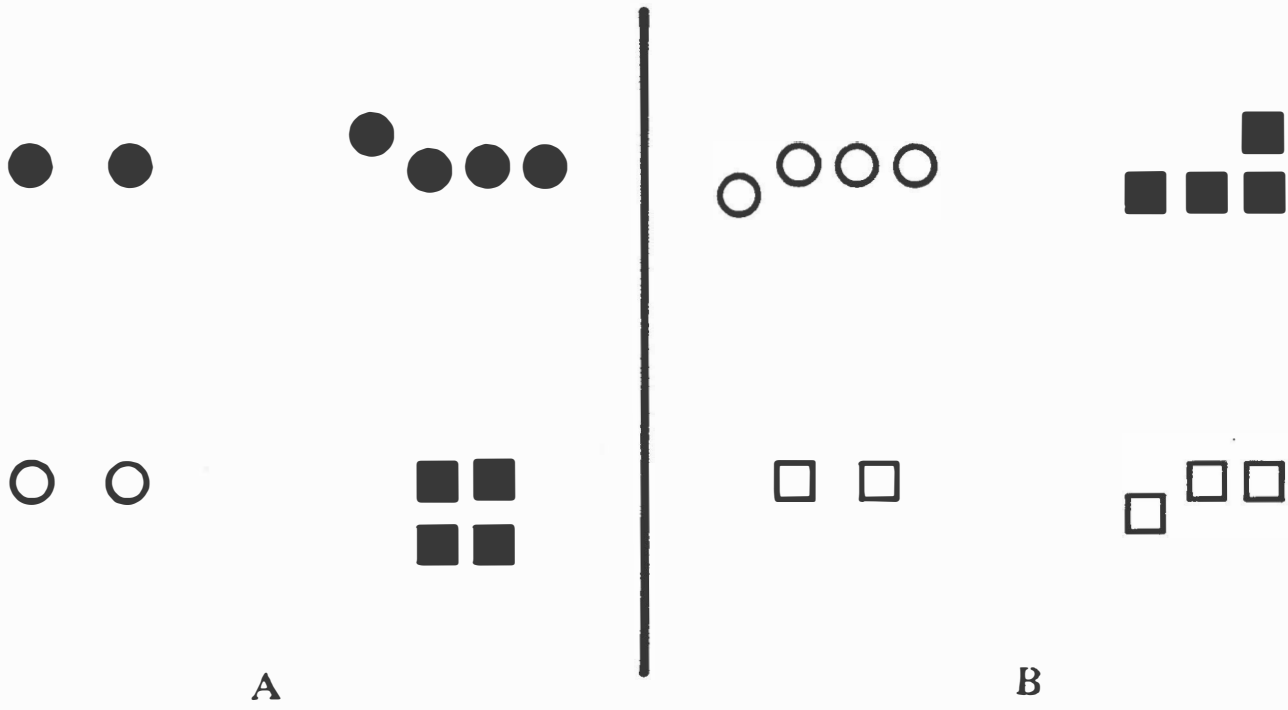

**2 OUT 3 POLYMORPHY:**

    **A: AT LEAST 2 OUT OF CIRCULAR, SYMMETRIC, AND BLACK**

    **B: AT LEAST 2 OUT OF SQUARE, ASYMMETRIC, AND WHITE**

**FIGURE 1**

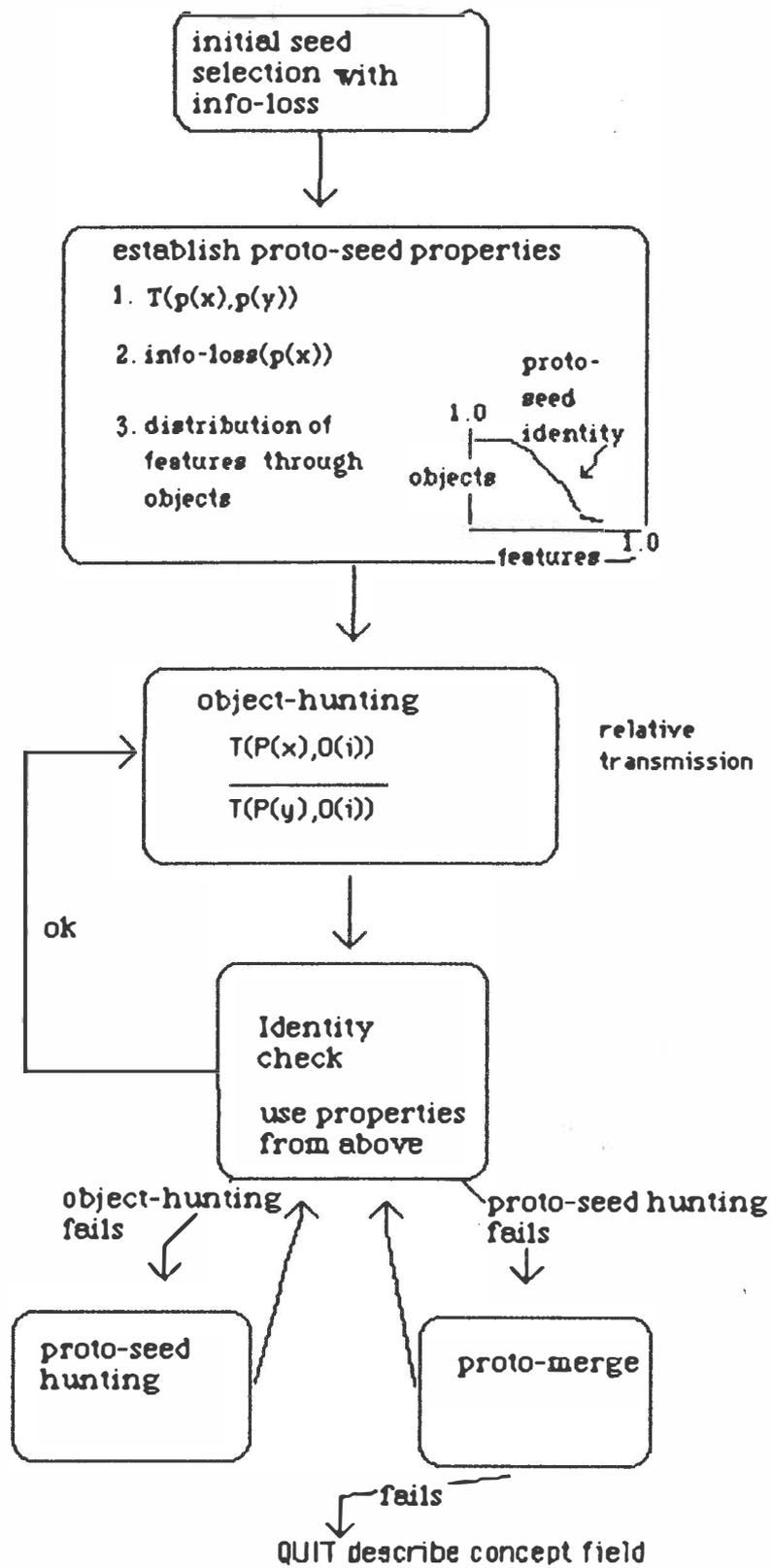

figure 2:    WITT flow

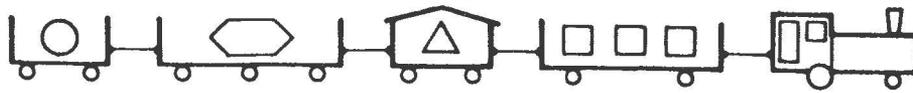
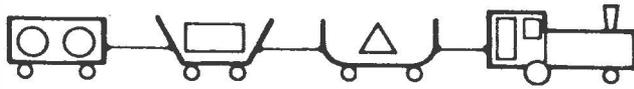
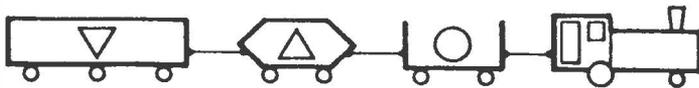
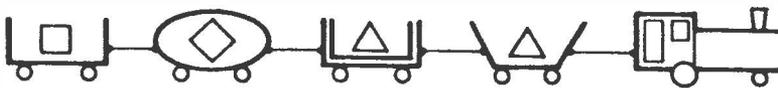
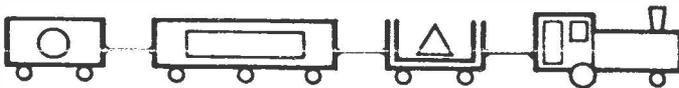

---

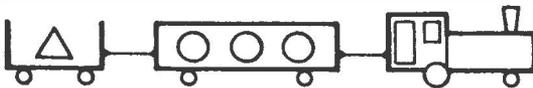
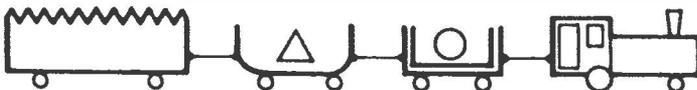
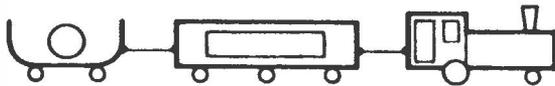
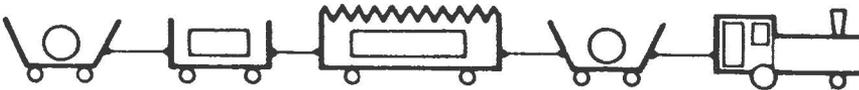
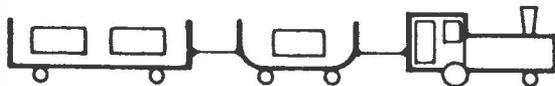

figure 3: "disjoint cover" example

```
%. PA V17 N 5 (1982) -- No. 52139        abstract 2
%A Weissinger-Baylon, Roger H.
%T Using models to solve problems: The functions of visual mental imagery. I: Text. II: Appendices.
%J Dissertation Abstracts International
%V 1979 Mar Vol 39(9-B) 4628-4629
%# 40550: PROBLEM SOLVING
%# 24470: IMAGERY

%. PA V17 N 5 (1982) -- No. 52149        abstract 1
%A Michaelis, Paul R.
%T Cooperative problem solving by like- and mixed-sex teams in a teletypewriter mode with unlimited, self-limited, introduced and
%J Dissertation Abstracts International
%V 1979 Mar Vol 39(9-B) 4632-4633
%# 21800: GROUP PROBLEM SOLVING
%# 23510: HUMAN SEX DIFFERENCES
%# 55520: VERBAL COMMUNICATION
%# 57230: WRITTEN LANGUAGE
%# 26250: INTERPERSONAL INTERACTION
%# 10970: COMPUTERS
%# 29350: MAN MACHINE SYSTEMS

%. PA V17 N 5 (1982) -- No. 52137        abstract 7
%A Viers, Gerald R.
%T Recognition and identification of visually presented words and pictures under shadowing conditions.
%J Dissertation Abstracts International
%V 1979 Mar Vol 39(9-B) 4628
%# 43350: RECOGNITION (LEARNING)
%# 38805: PICTORIAL STIMULI
%# 55575: VERBAL STIMULI
%# 55990: VISUAL STIMULATION
%# 23480: HUMAN INFORMATION STORAGE

%. PA V17 N 5 (1982) -- No. 52142        abstract 6
%A Yio, Jun H.
%T Visual recognition of words versus nonwords.
%J Dissertation Abstracts International
%V 1979 Mar Vol 39(9-B) 4630
%# 55987: VISUAL SEARCH
%# 57020: WORDS (PHONETIC UNITS)
%# 34340: NONSENSE SYLLABLE LEARNING
%# 24420: ILLUMINATION
%# 11560: CONTEXTUAL ASSOCIATIONS
%# 49220: SPELLING

%. PA V17 N 5 (1982) -- No. 52335        abstract 4
%A Richter, Gregory
%T The relationship between individual and developmental differences in scanning behavior and developmental trends in incidental
%J Dissertation Abstracts International
%V 1981 May Vol 41(11-B) 4287
%# 01360: AGE DIFFERENCES
%# 45540: SCHOOL AGE CHILDREN
%# 00950: ADOLESCENTS
%# 24700: INCIDENTAL LEARNING
%# 55987: VISUAL SEARCH

%. PA V17 N 5 (1982) -- No. 52499        abstract 3
%A Korant, Leslie L.
%T Effects of two visual training programs upon automaticity of letter and word recognition in urban Black kindergartners.
%J Dissertation Abstracts International
%V 1981 Jun Vol 41(12-A, Pt 1) 4959
%# 27370: KINDERGARTEN STUDENTS
%# 43350: RECOGNITION (LEARNING)
%# 57020: WORDS (PHONETIC UNITS)
%# 28230: LETTERS (ALPHABET)
%# 54940: URBAN ENVIRONMENTS
%# 06150: BLACKS
%# 16190: EDUCATIONAL PROGRAMS
%# 55980: VISUAL PERCEPTION

%. PA V17 N 5 (1982) -- No. 52613        abstract 5
%A Pushkash, Mark
%T Effect of the content of visually presented subliminal stimulation on semantic and figural learning task performance.
%J Dissertation Abstracts International
%V 1981 Jun Vol 41(12-A, Pt 1) 5036
%# 55550: VERBAL LEARNING
%# 34370: NONVERBAL LEARNING
%# 50470: SUBLIMINAL PERCEPTION
%# 55990: VISUAL STIMULATION
```

figure 4: psychological abstracts

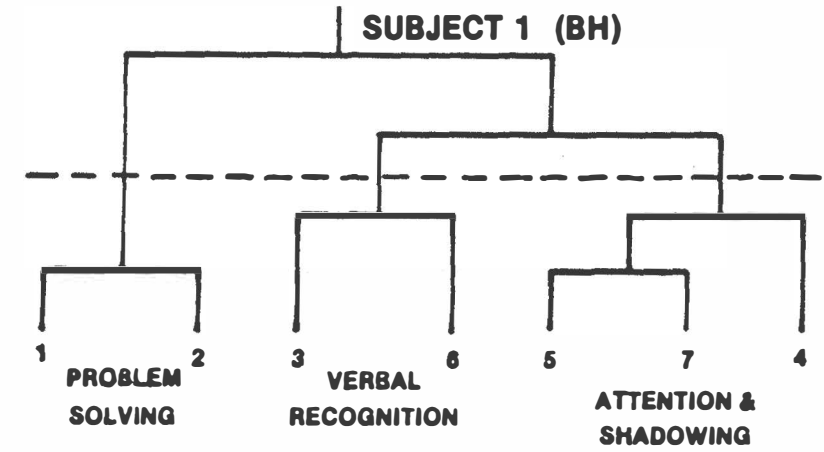
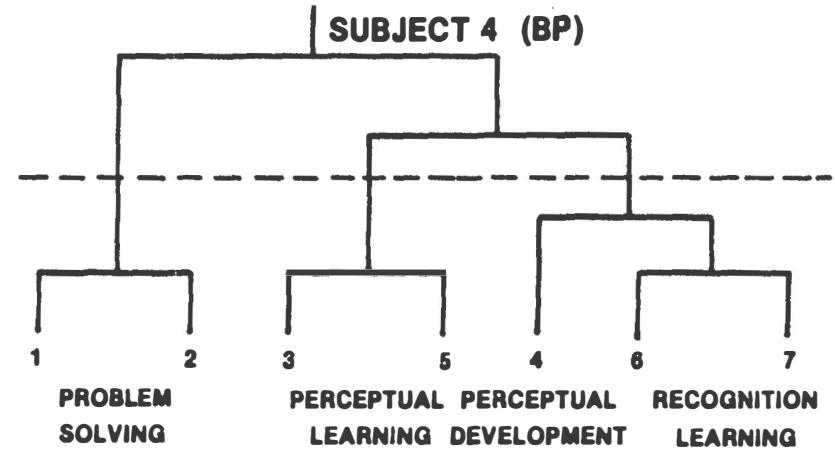
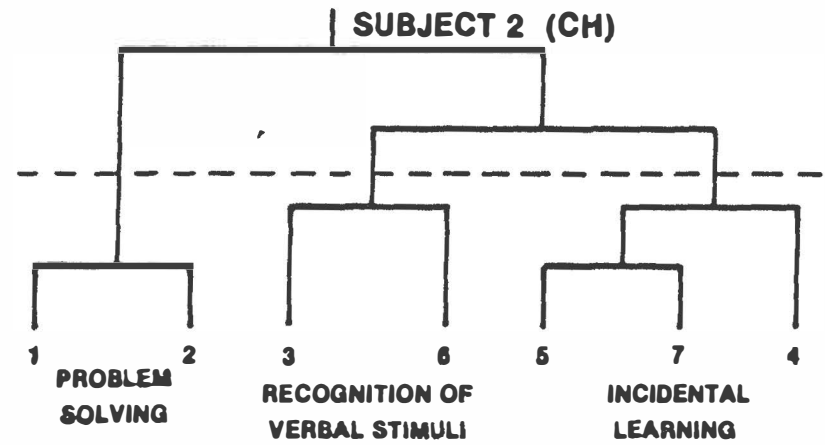
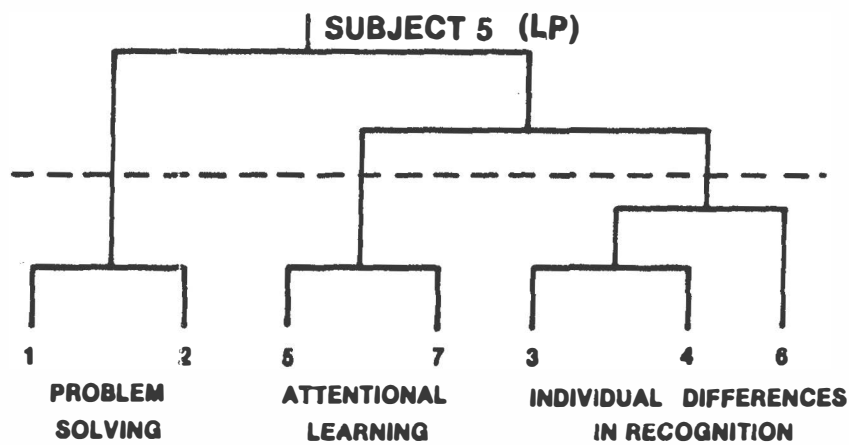
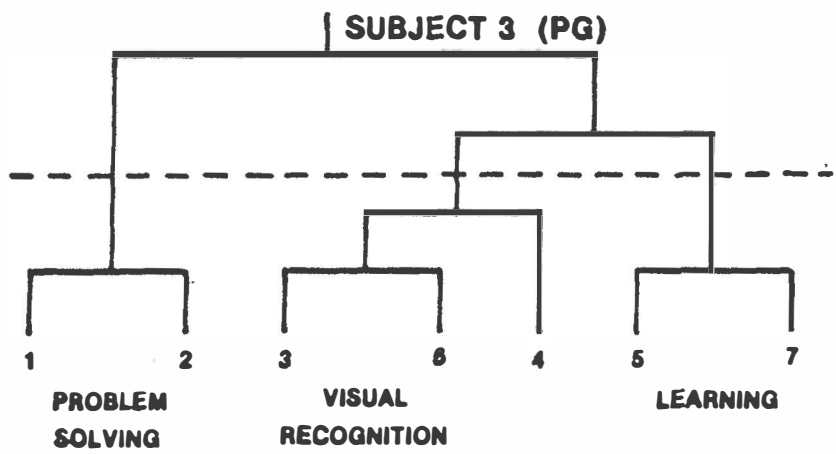
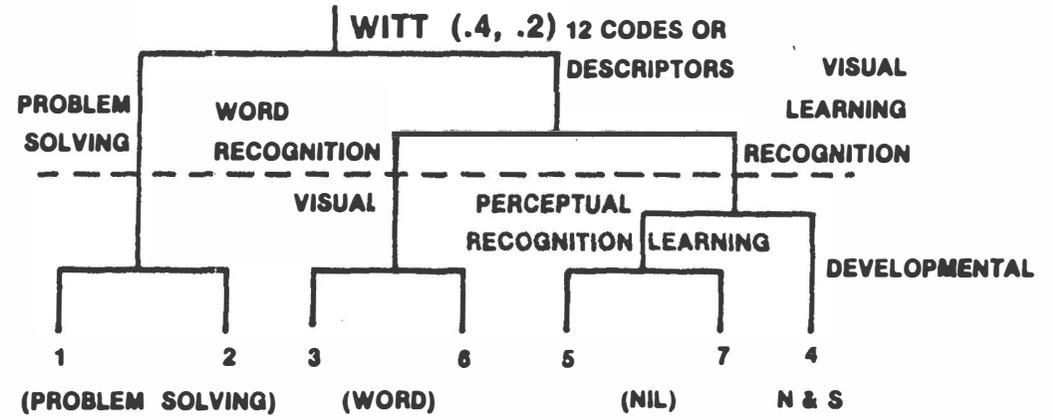

FIGURE 5